\documentclass[conference, compsoc]{IEEEtran} % IEEE Computer Society mode
\IEEEoverridecommandlockouts
\usepackage{cite}
\usepackage{amsmath,amssymb,amsfonts}
\usepackage{graphicx}
\usepackage{textcomp}
\usepackage{xcolor}

\usepackage{subcaption}
\usepackage{multirow}
\usepackage{booktabs}
\usepackage{algorithm, algpseudocode}
\usepackage{url}

\newcommand{\arxivcopyrightnotice}{%
    \copyright~2026 IEEE. Personal use of this material is permitted. Permission from IEEE must be obtained for all other uses, in any current or future media, including reprinting/republishing this material for advertising or promotional purposes, creating new collective works, for resale or redistribution to servers or lists, or reuse of any copyrighted component of this work in other works.
}

\newcommand{\del}[1]{}

\def\BibTeX{{\rm B\kern-.05em{\sc i\kern-.025em b}\kern-.08em
    T\kern-.1667em\lower.7ex\hbox{E}\kern-.125emX}}

\newcommand{\argmin}[1]{\underset{#1}{\mathrm{argmin}}\;}

\begin{document}

\title{TALLOP: Joint Optimization of Local Training and Communication Compression\\for Time-Constrained Federated Learning
\thanks{This work is supported in part by JST CREST JPMJCR21D2.\\Accepted for publication in The 32nd IEEE International Conference on Parallel and Distributed Systems (ICPADS 2026).\\\noindent\arxivcopyrightnotice\par}
}

\author{\IEEEauthorblockN{Honoka Anada}
\IEEEauthorblockA{\textit{The University of Tokyo / RIKEN}\\
Tokyo, Japan \\
honokaanada@is.s.u-tokyo.ac.jp}
\and
\IEEEauthorblockN{Yafei Ou}
\IEEEauthorblockA{\textit{RIKEN}\\
Tokyo, Japan \\
yafei.ou@riken.jp}
\and
\IEEEauthorblockN{Shinya Takamaeda-Yamazaki}
\IEEEauthorblockA{\textit{The University of Tokyo / RIKEN}\\
Tokyo, Japan \\
shinya@is.s.u-tokyo.ac.jp}
}

\maketitle

\begin{abstract}
Minimizing the wall-clock time required to reach a target accuracy is crucial in federated learning (FL), where many applications require frequent model updates as data distributions evolve over time. Nevertheless, in heterogeneous environments, slow clients with limited computational and communication capabilities can delay global synchronization, forcing faster clients to wait and increasing the overall training time. To mitigate this straggler effect, recent studies have introduced per-round time constraints and adapted client workloads within the available time. However, these approaches primarily focus on local computation, whereas communication can become a major bottleneck in communication-limited environments, motivating the joint optimization of training workload and communication compression. Although previous studies have explored such joint optimization, determining an effective training--communication balance under a fixed per-round time budget remains challenging, as the resulting model performance depends on both training progress and compression-induced degradation. We therefore propose TALLOP, which jointly optimizes the local training workload and communication compression under per-round time constraints. TALLOP predicts the training loss achievable under different computation budgets from the early-stage mini-batch loss trajectory (LENS), and estimates the degradation caused by communication compression using a compression-error model (CLEN). TALLOP combines these predictions to select the training-communication allocation that minimizes the post-compression loss under the given round time constraint. Experiments in heterogeneous FL environments show that TALLOP generally achieves competitive or improved time-to-accuracy compared with approaches that optimize only local training or neither, although the benefit over training-only adaptation depends on the setting.
\end{abstract}


\section{Introduction}
\begin{figure*}
    \centering
    \includegraphics[width=\linewidth]{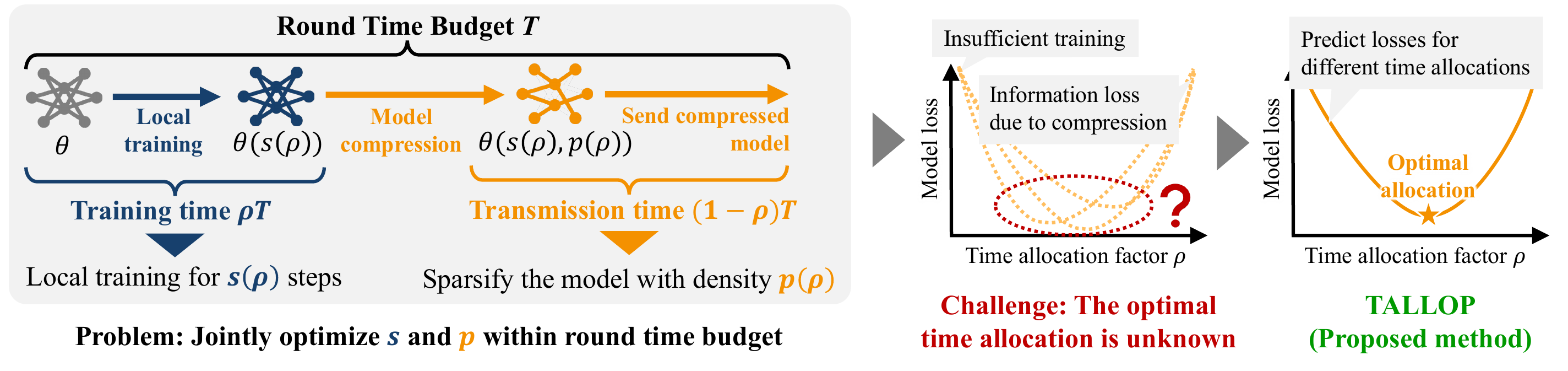}
    \caption{Given a fixed per-round time budget $T$, the time allocated to local training and communication determines the local training steps $s$ and the communication compression ratio $p$. While the optimal training-communication time allocation is unknown in advance, TALLOP predicts the post-compression model loss for different allocations and selects the one expected to achieve the best training outcome.}
    \label{fig:motivation}
\end{figure*}

Federated learning (FL)~\cite{mcmahan_communication-efficient_2017} enables collaborative model training while preserving data privacy by keeping training data on edge devices. In many practical applications, such as personalized recommendation~\cite{javeed_federated_2024} and keyboard prediction~\cite{hard_federated_2018}, data distributions continuously evolve over time, requiring frequent model updates. Therefore, reducing the wall-clock time to reach a target model accuracy is an important objective alongside merely improving accuracy.

This is particularly challenging under device heterogeneity, where differences in computational capability and communication bandwidth cause substantial variation in local training and model transmission time across clients. Under conventional synchronous FL, the duration of each communication round is determined by the slowest client, causing faster clients to remain idle while waiting for synchronization. This problem, known as the \textit{straggler effect}~\cite{pfeiffer_federated_2023}, can increase the time required to reach a target model accuracy.

To alleviate this problem, recent studies assign a fixed time budget to each communication round and adjust each client's training workload so that more clients can complete training within the deadline~\cite{shin_fedbalancer_2022, li_pyramidfl_2022, zhang_adaptive_2023, zhang_timelyfl_2023, narmadha_fedeff_2025}. However, these approaches primarily focus on adapting the local training workload and implicitly assume that communication occupies only a small fraction of each round. This assumption no longer holds in communication-limited environments, such as wireless edge networks, where transmitting model updates can take as long as, or even longer than, local training. In such environments, communication compression~\cite{konečný_federated_2016, lin_deep_2017} is as important as adaptive training workload for achieving time-efficient training.

Consequently, the per-round time budget creates a trade-off between local training and communication, as shown in Fig.~\ref{fig:motivation}. While allocating more time to training enhances the training quality, shorter communication time requires a higher compression ratio, resulting in greater information loss in model transmission. Conversely, allocating sufficient time for loss-free communication results in insufficient local training. Therefore, achieving better time-to-accuracy performance requires jointly allocating the available round time between training and communication according to the client's computation and communication capabilities.

Existing methods address only part of this problem. Deadline-aware FL methods optimize the local training workload, such as training steps~\cite{zhang_timelyfl_2023, zhang_adaptive_2023} or trained samples~\cite{shin_fedbalancer_2022}. However, they typically focus on computation, ignoring communication time. Meanwhile, some studies address the joint optimization of local training and communication compression in FL~\cite{nori_fast_2021, xu_adaptive_2023}. In particular, FedLamp~\cite{xu_adaptive_2023} optimizes local updating frequencies and compression ratios for heterogeneous clients based on convergence bounds. However, it does not directly optimize the training-communication time allocation under a fixed per-round time based on the predicted loss.

Consequently, \textit{prediction-based joint optimization of local training and communication compression under fixed per-round deadlines} remains insufficiently explored, particularly in heterogeneous and communication-limited environments. By predicting the loss resulting from different training-communication time allocations, such optimization could make more effective use of the available time budget, enabling greater learning progress per round and potentially improving time-to-accuracy.

In this paper, we propose \textbf{TALLOP}, \underline{T}raining-communication time \underline{AL}location via compressed model \underline{LO}ss \underline{P}rediction, which jointly determines the local training workload and communication compression ratio for each client under a given per-round time constraint. TALLOP predicts the model loss achievable under different training-communication time allocations and selects the allocation expected to minimize the post-compression loss. To enable this prediction, we introduce two key components. \textbf{LENS} predicts the model loss after future local training by extrapolating the mini-batch loss trajectory, while \textbf{CLEN} estimates the additional loss caused by communication compression using a lightweight compression-error model. By combining these two components, TALLOP efficiently evaluates candidate training-communication time allocations during local training. Experiments demonstrate that (1) jointly optimizing the training workload and communication compression generally provides competitive or improved time-to-accuracy compared with optimizing either or neither, and (2) the computational overhead of TALLOP is equivalent to the cost of training on only a few samples, resulting in a negligible increase in the overall training time.

\section{Related Work}

\subsection{Time-to-Accuracy Optimization under Device Heterogeneity in FL}

In FL, improving time-to-accuracy under device heterogeneity is challenging because a small number of stragglers can significantly delay global synchronization~\cite{pfeiffer_federated_2023}. Existing approaches can be broadly classified into two categories.

One approach is asynchronous FL~\cite{xie_asynchronous_2019, ma_fedsa_2021, nguyen_federated_2022}, which removes synchronization barriers by allowing clients to upload model updates independently. This alleviates the straggler problem but introduces the \textit{stale update} problem, in which outdated model updates may degrade convergence. Moreover, it is incompatible with mechanisms that rely on synchronous rounds, such as secure aggregation.

In contrast, another line of work retains synchronous aggregation while adapting the local training workload according to each client's computational capability. By assigning more local computation to fast clients and less to slow ones, these methods reduce idle waiting time while preventing slow devices from becoming stragglers. This idea has been explored by adapting various aspects of local training, including the number of local epochs~\cite{reisizadeh_fedpaq_2020, nguyen_efficient_2025}, batch sizes~\cite{ma_adaptive_2023, park_amble_2022}, and parameters updated during local training~\cite{li_fedsae_2021, liu_adaptivefl_2025}.

This idea can be further enhanced by introducing a fixed deadline for each communication round~\cite{nishio_client_2019} and adapting the local workload so that clients can complete their computation and model upload within the available time. FedBalancer~\cite{shin_fedbalancer_2022}, for example, jointly determines the round deadline and training sample selection, while PyramidFL~\cite{li_pyramidfl_2022}, TimelyFL~\cite{zhang_timelyfl_2023}, and FedState~\cite{zhang_adaptive_2023} personalize the amount of local computation, such as training iterations, epochs, updated parameters, or training samples. These methods improve time-to-accuracy by making better use of heterogeneous computation capabilities, but generally focus on local computation and do not explicitly optimize communication time.

This limitation becomes important in communication-limited environments, where model transmission can occupy a substantial fraction of the round time. In such settings, the communication compression ratio also determines how much local training can be performed within a fixed time budget: stronger compression can shorten communication but may introduce greater information loss. Thus, training workload and communication compression are coupled through the per-round time constraint, motivating approaches that jointly optimize the two.

\subsection{Joint Optimization of Training and Communication in FL}

Meanwhile, several studies have jointly optimized the local training workload and communication cost to improve training efficiency. For example, Wang and Joshi~\cite{wang_communication_2019, wang_federated_2019} optimized the model synchronization interval, while Han \textit{et al.}~\cite{han_adaptive_2020} adaptively adjusted the communication compression ratio to minimize training time. AdaQuantFL~\cite{jhunjhunwala_adaptive_2021} optimized the quantization level based on convergence analysis under a communication budget, and FFL~\cite{nori_fast_2021} jointly optimized the local training workload and communication sparsity through theoretical convergence analysis.

Most closely related to our work, Xu \textit{et al.} proposed FedLamp~\cite{xu_adaptive_2023}, which jointly optimizes local updating frequencies and model compression ratios for heterogeneous clients based on a theoretical convergence bound.

However, convergence-bound-based optimization does not directly estimate the actual learning progress achieved by different training-communication allocations within a fixed round deadline. In heterogeneous and communication-limited environments, the effectiveness of each allocation can vary across clients and training rounds, depending on both system capabilities and the current learning behavior. Therefore, a prediction-based approach that estimates the loss after local training and compression is needed to select an effective allocation under a given time budget. Such an approach could better utilize the available round time and improve time-to-accuracy.

\section{Methodology}

\subsection{Problem Formulation}

In this section, we formulate the optimization problem addressed in this paper. We first describe the assumed FL system model and then present the optimization objective.

\subsubsection{System Model}

Let $\mathcal{N} := \{1,2,\dots,N\}$ denote the set of all clients participating in FL. Each client $k \in \mathcal{N}$ possesses its own local training dataset $\mathcal{D}_k$. The global model parameters are updated synchronously in each communication round. Each round $r$ is assigned a time budget $T^{(r)}$, and updates arriving after the deadline are excluded from aggregation.

Let $\theta^{(r)}$ denote the global model parameters at the beginning of round $r$. At the beginning of each round, the central server randomly selects $K$ participating clients and distributes $\theta^{(r)}$ to them. Each selected client $k$ performs $s_k^{(r)}$ steps of mini-batch training with batch size $B$, obtaining the locally trained parameters $\theta_k^{(r)}(s_k^{(r)})$ as follows:
\begin{equation}
\begin{aligned}
\theta_k^{(r)}(0) &= \theta^{(r)},\\
\theta_k^{(r)}(t+1) &= \theta_k^{(r)}(t)
-\eta\nabla l\!\left(\theta_k^{(r)}(t);\mathcal{B}_t\right),\\
&\quad\quad\quad\quad\quad t=0,\ldots,s_k^{(r)}-1.
\end{aligned}
\end{equation}
where $\mathcal{B}_t$ denotes the mini-batch used at the $t$-th training step. The client then compresses the parameter difference before uploading the compressed update to the central server. While various update compression methods have been proposed, including quantization~\cite{wen_terngrad_2017, shlezinger_uveqfed_2021} and pruning~\cite{lin_deep_2017, li_lotteryfl_2020}, we focus on \textit{sparsification with unstructured pruning} and leave other compression methods out of scope. Specifically, only the top $p_k^{(r)}$ fraction of elements with the largest absolute values in the parameter difference vector are transmitted, as follows:
\begin{equation}
\label{eq:sparsification}
\theta_k^{(r)}(s_k^{(r)}, p_k^{(r)})=\theta^{(r)}+\mathcal{S}\left(\theta_k^{(r)}(s_k^{(r)})-\theta^{(r)};\;p_k^{(r)}\right),
\end{equation}
where $\mathcal{S}(\cdot;p)$ denotes the sparsification operator that retains only the top $p$ fraction of elements with the largest absolute values. The central server then aggregates the received updates to generate the global model for the next round. We further assume that each transmitted parameter element is represented using 32 bits. While this study focuses on sparsification, extending the optimization to other communication compression methods is left for future work.

Hereafter, client $k$ and round $r$ are assumed to be fixed, and the subscript $k$ and superscript $(r)$ are omitted whenever there is no ambiguity.

In this study, we assume that model downloads are performed without compression, and we do not consider their optimization, denoting the download time as $T_\textit{down}$. This is a scope choice: the proposed optimization concerns local training and upload compression only. The model training and upload times are uniquely determined by the local training step count $s$ and the transmission ratio $p$, respectively, denoted by $T_\textit{train}(s)$ and $T_\textit{up}(p)$.

\subsubsection{Optimization Objective}

The local training workload and transmission ratio jointly affect model quality, while competing for the same per-round time budget. Our objective is therefore to jointly determine the local training steps $s$ and the transmission ratio $p$ to minimize the loss of the transmitted model while satisfying the per-round time constraint. Let $l(\theta)$ denote the loss of model $\theta$ evaluated on the local dataset $\mathcal{D}_k$. The optimization problem is formulated as follows:

\begin{equation}
    \label{eq:objective}
    \argmin{s \in \mathbb{N}_{\ge 0}, p \in [0, 1]} l(\theta(s, p)) \quad\text{s.t.}\quad T_\textit{down} + T_\textit{train}(s) + T_\textit{up}(p) \le T
\end{equation}

\subsubsection{Time Allocation Parameter}

Directly optimizing two coupled variables subject to a time constraint is inconvenient. We therefore reparameterize the feasible solution space by introducing a scalar \textit{time-allocation parameter} $\rho \in [0,1]$, representing the fraction of the available round time allocated to local training after the model download.

Let $T' := T - T_\textit{down}$ be the remaining time available for local computation and model upload. Given a time allocation $\rho$, the client performs the maximum possible number of local training steps within the allocated training time $\rho T'$, and then transmits the model using the maximum feasible transmission ratio that satisfies the remaining communication time $(1-\rho)T'$. Hence, every feasible value of $\rho$ uniquely determines a pair of local training steps $s(\rho)$ and transmission ratio $p(\rho)$. Consequently, the joint optimization problem can be reformulated as a \textit{per-round time allocation problem between local training and model transmission}, as follows.

\begin{equation}
    \label{eq:objective-rho}
    \argmin{\rho \in [0, 1]} l(\theta(s(\rho), p(\rho)))
\end{equation}

\subsubsection{Mapping Time Allocation to Local Training and Communication}

Given the time allocation ratio $\rho$, the local training steps $s$ and the transmission ratio $p$ are uniquely determined according to the client's computational capability and communication bandwidth.

Let the computational performance of client $k$ be $F$ FLOPS, and let the download and upload bandwidths be $\beta_{\textit{down}}$ and $\beta_{\textit{up}}$ bps, respectively. Let $|\theta|$ denote the number of model parameters, $M$ the number of floating-point operations required to process one training sample, and $B$ the mini-batch size. We assume that the training time is proportional to the number of floating-point operations, while other bottlenecks, such as memory bandwidth, are ignored.

The model download time is given by
\begin{equation}
T_\textit{down}=\frac{32|\theta|}{\beta_{\textit{down}}},
\end{equation}
and the remaining time $T' = T - T_\textit{down}$ is divided between local training and model upload. Since the computation time required for one local training step is approximated by $\frac{MB}{F}$, the maximum number of local training steps that can be performed is given by 
\begin{equation}
s(\rho)=\left\lfloor\frac{F}{MB}\rho T'\right\rfloor.
\end{equation}

Meanwhile, the transmission ratio $p$ can be determined by the remaining time after local training, $(1-\rho)T'$. When $p<1$, the update is transmitted in a sparse vector format consisting of pairs of non-zero values and their corresponding indices. Since the number of bits required to represent each index is $8\cdot\left\lceil\log_2(|\theta|) / 8\right\rceil$, the total transmission size is
\begin{equation}
\left(32+8\left\lceil\frac{\log_2(|\theta|)}{8}\right\rceil\right)p|\theta|.
\end{equation}
Hence, the total upload time is given by
\begin{equation}
T_\textit{up}(p):=
\begin{cases}
    \frac{\left(32+8\left\lceil\frac{\log_2(|\theta|)}{8}\right\rceil\right)p|\theta|}{\beta_\textit{up}} & \text{if $p < 1$}, \\
    \frac{32|\theta|}{\beta_\textit{up}} & \text{if $p = 1$}.
\end{cases}
\end{equation}
Consequently, the maximum feasible transmission ratio $p(\rho)$ is given by
\begin{equation}
\label{eq:p-rho}
p(\rho) =
\begin{cases}
1,
& \text{if } \beta_{\mathrm{up}}(1-\rho)T' \geq 32|\theta|, \\[6pt]
\displaystyle
\frac{\beta_{\mathrm{up}}(1-\rho)T'}
{\left(32+8\left\lceil\frac{\log_2(|\theta|)}{8}\right\rceil\right)|\theta|},
& \text{otherwise}.
\end{cases}
\end{equation}

\subsection{TALLOP Design}

\begin{figure*}
    \centering
    \includegraphics[width=0.8\linewidth]{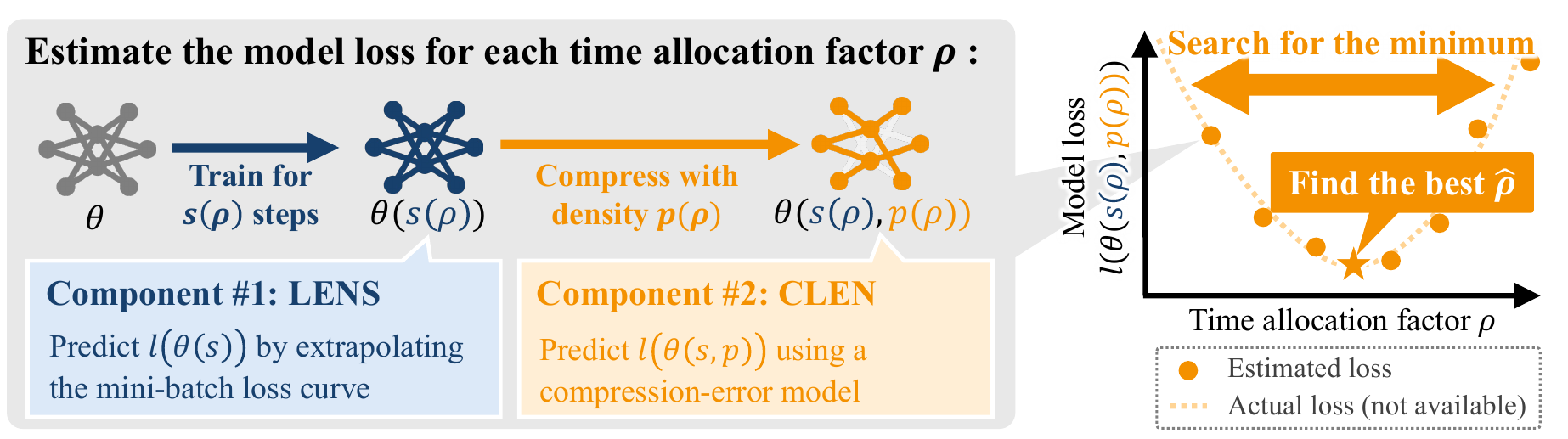}
    \caption{Overview of TALLOP. TALLOP predicts the post-compression loss by combining LENS and CLEN, and determines the optimal time allocation factor $\rho$ by minimizing the predicted loss.}
    \label{fig:tallop}
\end{figure*}

A straightforward approach to solve Eq.~\eqref{eq:objective-rho} is to evaluate the loss after local training and compression, $l(\theta(s(\rho), p(\rho)))$, for multiple candidate values of $\rho$ and identify the minimizer. However, this approach faces two fundamental challenges. \textit{First}, exact evaluation of $l(\theta(s(\rho), p(\rho)))$ requires computing the loss for multiple candidate values of $\rho$. Since each evaluation requires model inference on a sufficiently large number of samples, repeating this process incurs substantial computational overhead, which contradicts the objective of improving time efficiency. \textit{Second}, at the current training step $s_0$, the model loss corresponding to a future training step $s>s_0$ is unknown, making it impossible to precisely determine the optimal training termination time.

Given these challenges, we propose TALLOP, which determines the optimal $\rho$ by \textit{predicting} the post-compression model loss $l(\theta(s(\rho), p(\rho)))$ at the current and future training steps. TALLOP consists of the following two components, as shown in Fig.~\ref{fig:tallop}:

\begin{itemize}
    \item \underline{L}oss-curve \underline{E}xtrapolatio\underline{N} via \underline{S}moothing (\textbf{LENS}): predicts the post-training future loss $l(\theta(s))$ by smoothing and extrapolating the mini-batch loss trajectory.
    \item \underline{C}ompressed model \underline{L}oss estimation by \underline{E}rror \underline{N}orm (\textbf{CLEN}): estimates the post-compression model loss $l(\theta(s, p))$ from the norm of the compression-induced parameter error.
\end{itemize}

By combining these two components, TALLOP constructs an estimate of the post-compression model loss for any time allocation factor $\rho$ as
\begin{equation}
\tilde l(\rho) \approx l(\theta(s(\rho), p(\rho))).
\end{equation}
The optimal time allocation factor is then obtained by minimizing the predicted loss using ternary search:
\begin{equation}
\hat\rho=\argmin{\rho\in[0,1]}\tilde l(\rho).
\end{equation}

Finally, the client continues local training until $s(\hat\rho)$ training steps have been completed, compresses the resulting model update using the transmission ratio $p(\hat\rho)$, and transmits the compressed update to the server. The remainder of this section first describes LENS and CLEN in detail, followed by the overall optimization algorithm.

\subsubsection{LENS: Loss-Curve Extrapolation via Smoothing}

This part describes how to estimate $l(\theta(s))$ for an arbitrary training step $s$. Since future training steps have not yet been executed, their corresponding losses cannot be directly observed. Moreover, evaluating the loss on a validation dataset at every completed training step incurs prohibitive computational overhead.

Instead, the proposed method estimates the future loss using the history of mini-batch losses observed up to the current training step $s_0$:
\begin{equation}
\label{eq:minibatch-loss-history}
\left\{l\left(\theta(s);\,\mathcal{B}_s\right) \;|\; s \in [1, s_0]  \right\}.
\end{equation}
However, directly using these observations is inappropriate because the mini-batch losses contain substantial batch-dependent noise and provide no information about future steps ($s > s_0$).

To address this issue, we model the underlying loss curve as a monotonically decreasing function that asymptotically converges during training. Specifically, the loss curve is represented by
\begin{equation}
\label{eq:loss-predict-model}
f_\textit{loss}(s;\,l_\textit{inf}, a, b, c) = l_\textit{inf} + \frac{a}{(s+b)^c},
\end{equation}
where the parameters $l_\textit{inf}$, $a$, $b$, and $c$ are estimated so that the function best fits the observed mini-batch loss sequence. 

By fitting this model, LENS both smooths the noisy mini-batch loss trajectory and extrapolates the trajectory to unobserved training steps. The predicted loss at an arbitrary step $s$ is therefore given by
\begin{equation}
\label{eq:predict-future-loss}
\tilde{l}(\theta(s)) := f_\textit{loss}(s;\,\hat{l}_\textit{inf}, \hat{a}, \hat{b}, \hat{c}) \approx l(\theta(s)),
\end{equation}
where $\hat{l}_\textit{inf}$, $\hat{a}$, $\hat{b}$, and $\hat{c}$ denote the estimated model parameters.

We fit the loss curve using bounded nonlinear least squares, with constraints $l_{\mathrm{inf}},a,c\geq0$, $b\geq10^{-6}$, and $c\leq5$. The parameters are initialized as $(l_{\mathrm{inf}},a,b,c)=(0.1l_{\min},\,l_1-0.1l_{\min},\,1,\,0.5)$, where $l_{\min}$ and $l_1$ are the minimum and first observed losses, respectively. If fewer than three observations are available or fitting fails to converge, the initial parameters are used as a fallback. This fitting assumes a smooth, monotonically decreasing loss trajectory, which may not hold for short or noisy local training sequences.

\subsubsection{CLEN: Compressed-model Loss Estimation by Error Norm}

LENS provides estimates of the model losses before and after local training, namely $\tilde{l}(\theta)=\tilde{l}(\theta(0))$ and $\tilde{l}(\theta(s))$. We next estimate the post-compression loss $l(\theta(s,p))$ using these estimates. To this end, we first establish a lightweight compression-loss model based on theoretical insight, and then derive a post-compression loss estimator from it. Finally, we empirically validate the effectiveness of the proposed compression-loss model.

\textbf{Lightweight Compression-Loss Model.}
Consider a trained model $\theta$ perturbed by a small parameter perturbation $\Delta\theta$. Assuming that the loss function $l(\cdot)$ is twice differentiable in the neighborhood of $\theta$, a second-order Taylor expansion yields
\begin{equation}
l(\theta+\Delta\theta)=l(\theta)+\nabla l(\theta)^\top \Delta\theta+\frac{1}{2}\Delta\theta^\top H(\theta)\Delta\theta+o(\lVert\Delta\theta\rVert_2^2)
\end{equation}
where $H(\theta)$ denotes the Hessian matrix. Since a trained model approximately satisfies $\nabla l(\theta)\approx 0$,
\begin{equation}
\Delta l:=l(\theta+\Delta\theta)-l(\theta)\approx\frac{1}{2}\Delta\theta^\top H(\theta)\Delta\theta
\end{equation}
holds. Furthermore, when the perturbation does not exhibit a strong directional bias, the expected loss increase can be approximated as
\begin{equation}
\label{eq:loss_delta_theta_relation}
\Delta l\propto\lVert\Delta\theta\rVert_2^2.
\end{equation}
Meanwhile, these assumptions are limitations of CLEN: $\nabla l(\theta)\approx 0$ is most appropriate near a stationary point and may not hold in the early stages of training, while the absence of strong directional bias is also a strong assumption. Consequently, the accuracy of CLEN can depend on the architecture, layer, and training stage.

\textbf{Post-Compression Loss Estimator. }
Next, we establish a post-compression loss estimator based on this observation. Let the com\-pres\-sion-induced perturbation after $s$ local training steps at client $k$ be defined as
\begin{equation}
\Delta \theta(s,p)=\theta(s,p)-\theta(s).
\end{equation}
Using Eq.~\eqref{eq:loss_delta_theta_relation}, the increase in loss caused by compression is approximated as
\begin{equation}
l(\theta(s,p))-l(\theta(s))\approx\lambda\lVert\Delta \theta(s,p)\rVert_2^2,
\end{equation}
where $\lambda>0$ is a proportionality constant. Since $\theta(s,0)=\theta$,
\begin{equation}
\lambda=\frac{l(\theta)-l(\theta(s))}{\lVert\theta(s)-\theta\rVert_2^2}
\end{equation}
can be estimated.

Accordingly, the compressed-model loss is approximated as
\begin{equation}
\label{eq:predict-compressed-loss}
\tilde{l}(\theta(s,p)):=\tilde{l}(\theta(s))+\frac{\tilde{l}(\theta)-\tilde{l}(\theta(s))}{\lVert\theta(s)-\theta\rVert_2^2}\lVert\theta(s,p)-\theta(s)\rVert_2^2.
\end{equation}

\textbf{Empirical Validation. }
Finally, we experimentally validate the accuracy of this approximation using two representative neural network architectures: MobileNetV2~\cite{sandler_mobilenetv2_2018} and Vision Transformer (ViT)~\cite{dosovitskiy_image_2020}. For each model, let $\theta_0$ denote the initial parameters and $\theta$ denote the parameters obtained after one epoch of training on CIFAR-10~\cite{cifar10}. A sparsified model with non-zero ratio $p$ is then constructed according to Eq.~\eqref{eq:sparsification} as
\begin{equation}
    \theta(p) = \theta_0 + \mathcal{S}(\theta - \theta_0;\;p).
\end{equation}

We compare the loss predicted by Eq.~\eqref{eq:predict-compressed-loss} with the actual loss obtained by directly evaluating $\theta(p)$. To verify that the approximation remains valid beyond the initial training stage, we additionally repeat the same experiment by using the parameters after two training epochs (\texttt{Epoch 2}) as $\theta_0$.

\begin{figure}[t]
    \centering
    \begin{subfigure}{0.48\linewidth}
        \centering
        \includegraphics[width=\linewidth]{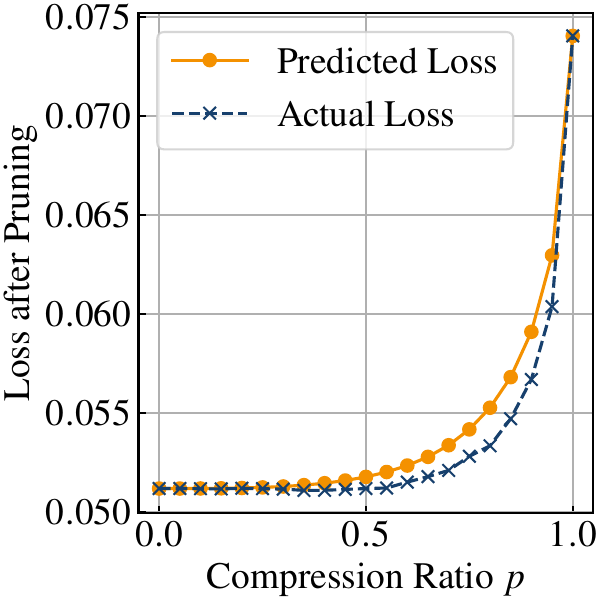}
        \caption{MobileNetV2 (Epoch 0)}
    \end{subfigure}
    \begin{subfigure}{0.48\linewidth}
        \centering
        \includegraphics[width=\linewidth]{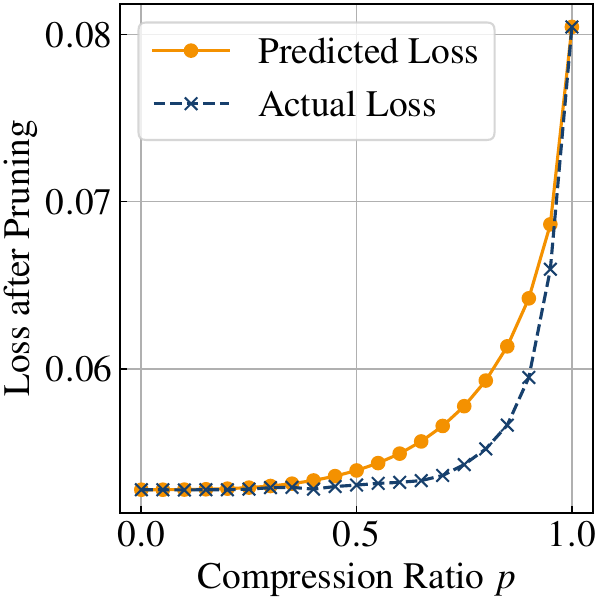}
        \caption{ViT (Epoch 0)}
    \end{subfigure}
    \begin{subfigure}{0.48\linewidth}
        \centering
        \includegraphics[width=\linewidth]{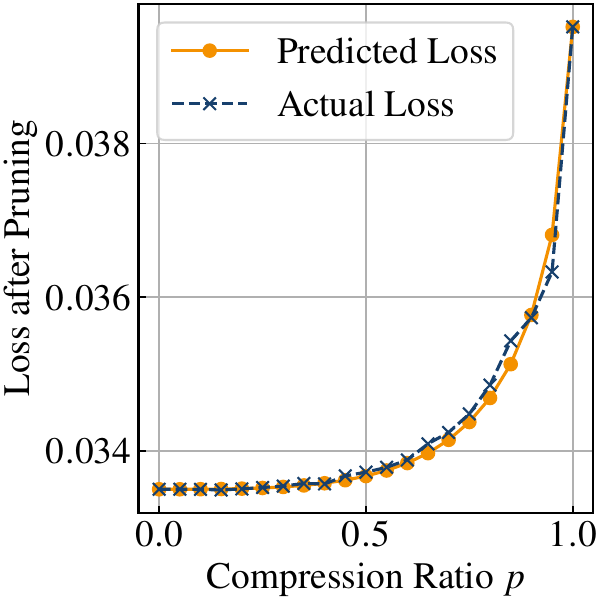}
        \caption{MobileNetV2 (Epoch 2)}
    \end{subfigure}
    \begin{subfigure}{0.48\linewidth}
        \centering
        \includegraphics[width=\linewidth]{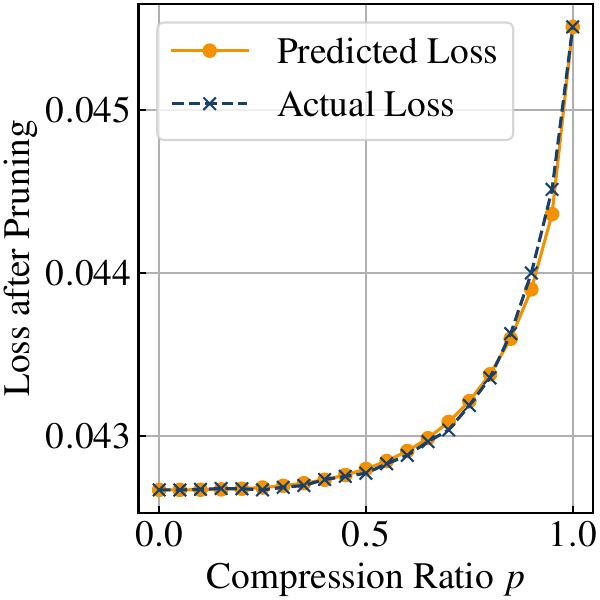}
        \caption{ViT (Epoch 2)}
    \end{subfigure}
    \caption{Actual and predicted losses of sparsified models for various non-zero ratios $p$. The proposed CLEN accurately estimates the loss after compression.}
    \label{fig:predict-compressed-loss-result}
\end{figure}

Figure~\ref{fig:predict-compressed-loss-result} compares the predicted and actual losses of the compressed models for various non-zero ratios $p$. The predicted losses closely match the measured losses across both network architectures and training stages, validating the effectiveness of CLEN for estimating the loss after compression. A slight tendency to overestimate the loss can be observed under the \texttt{Epoch~0} setting. Improving the estimation accuracy during the earliest stages of training remains an interesting direction for future work.

\subsubsection{Overall Algorithm of TALLOP}

Finally, TALLOP combines LENS and CLEN to determine the optimal time allocation factor $\rho$ in each communication round. Given a candidate value of $\rho$, LENS predicts the post-training loss $\tilde{l}(\theta(s(\rho)))$, CLEN estimates the corresponding post-compression loss $\tilde{l}(\theta(s(\rho),p(\rho))) =: \tilde{l}(\rho)$, and the optimal allocation factor $\hat{\rho}$ is obtained by minimizing the predicted compressed-model loss. Assuming that $\tilde{l}(\rho)$ is unimodal with respect to $\rho$, TALLOP identifies $\hat{\rho}$ by ternary search. The reliance on this unimodality assumption is a limitation; optimization strategies that do not require it are left for future work.

A practical implementation of this procedure must address two issues. \textit{First}, when should the prediction be executed during local training? \textit{Second}, how can the predicted compressed-model loss be computed efficiently using only the information available at the current training step? The remainder of this section addresses these two issues.

\textbf{Estimation timing. }
If prediction cost is ignored, the highest prediction accuracy is achieved by performing the above prediction after every local training step and terminating local training once the predicted optimal number of steps has been reached. However, repeatedly performing this prediction incurs non-negligible computational overhead, offsetting the time savings gained through optimized workload allocation.

Instead, TALLOP performs the prediction only once during each communication round. Let $s_0$ denote the maximum number of local training steps that still allows the client to upload the model without compression:

\begin{align}
s_0 &= \max_{s \in \mathbb{Z}_{\ge 0}} s
\quad\text{s.t.}\quad
T_\textit{train}(s) + T_\textit{up}(1) \le T'
\end{align}

By construction, any training workload satisfying $s \le s_0$ can upload an uncompressed model update and therefore introduces no compression error. Assuming that the model loss decreases monotonically with the number of local training steps, the objective value also decreases monotonically for all $\rho$ satisfying $s(\rho)\le s_0$. Consequently, the optimal allocation factor must satisfy $s(\rho)\ge s_0$. Therefore, the prediction is performed once, at the first mini-batch boundary within one batch of $s_0$, or earlier if the maximum local training workload is reached. After estimating the optimal training workload $\hat{s}$, TALLOP selects the first mini-batch boundary at which the next batch would reach or exceed $\hat{s}$.

\textbf{Practical loss estimation. }
Another practical challenge arises when evaluating Eq.~\eqref{eq:predict-compressed-loss}. The prediction requires the future model parameters $\theta(s)$ and their compressed counterpart $\theta(s,p)$ for arbitrary training steps $s \ge s_0$, which are unavailable at the estimation point $s_0$.

To address this issue, TALLOP approximates the required parameter differences using the model obtained at the estimation point $s_0$. Specifically, $\theta(s)$ and $\theta(s,p)$ are replaced by $\theta(s_0)$ and $\theta(s_0,p)$, yielding the following practical estimator:
\begin{equation}
\label{eq:predict-compressed-loss-feasible}
\tilde{l}'(\theta(s,p)):=\tilde{l}(\theta(s))+\frac{\tilde{l}(\theta)-\tilde{l}(\theta(s))}{\lVert\theta(s_0)-\theta\rVert_2^2}\lVert\theta(s_0,p)-\theta(s_0)\rVert_2^2.
\end{equation}
In practice, $\tilde{l}(\theta)$ is approximated by the maximum observed mini-batch loss to avoid unstable extrapolation to $s=0$.

\textbf{Overall procedure. }
Finally, TALLOP performs ternary search using the above estimator to determine the optimal time allocation factor $\hat{\rho}$, from which the corresponding local training workload $s(\hat{\rho})$ and transmission ratio $p(\hat{\rho})$ are obtained. The complete procedure is summarized in Algorithm~\ref{alg:proposal}.

\begin{algorithm}[!t]
    \caption{Procedure to determine the local training steps $s$ and transmission ratio $p$ by TALLOP}\label{alg:proposal}
    \begin{algorithmic}[1]
        \Function{PredictCompressedLoss}{$\rho$}
            \State $\tilde{l}(\theta) \gets f_\textit{loss}(0; \hat{l}_\textit{inf}, \hat{a}, \hat{b}, \hat{c})$
            \State $\tilde{l}(\theta(s)) \gets f_\textit{loss}(s; \hat{l}_\textit{inf}, \hat{a}, \hat{b}, \hat{c})$
            \State $\theta(s_0,p(\rho)) \gets \theta+\mathcal{S}\left(\theta(s_0)-\theta;\;p(\rho)\right)$
            \State Calculate $\tilde{l}'(\theta(s(\rho),p(\rho)))$ based on Eq.~\eqref{eq:predict-compressed-loss-feasible}
            \State \Return $\tilde{l}'(\theta(s(\rho),p(\rho)))$
        \EndFunction
        \vspace{5mm}
        \Function{TALLOP}{$\mathcal{L}_\textit{history}$}
            \State $\hat{l}_\textit{inf}, \hat{a}, \hat{b}, \hat{c} \gets $ Fit past minibatch losses $\mathcal{L}_\textit{history}$ with Eq.~\eqref{eq:loss-predict-model}
            \State $\rho_\textit{min} \gets 0,\quad \rho_\textit{max} \gets 1$
            \While{$\rho_\textit{max} - \rho_\textit{min} > \varepsilon$}
                \State $\rho_1 \gets \frac{2 \rho_\textit{min} + \rho_\textit{max}}{3}$
                \State $\rho_2 \gets \frac{\rho_\textit{min} + 2 \rho_\textit{max}}{3}$
                \State $l_1 \gets \Call{PredictCompressedLoss}{\rho_1}$
                \State $l_2 \gets \Call{PredictCompressedLoss}{\rho_2}$
                \If {$l_1 < l_2$}
                    \State $\rho_\textit{max} \gets \rho_2$
                \Else
                    \State $\rho_\textit{min} \gets \rho_1$
                \EndIf
            \EndWhile
            \State \Return $s(\rho_\textit{min}),\;p(\rho_\textit{min})$
        \EndFunction
        \vspace{5mm}
        \Function{TrainAndUpload}{}
            \State $\theta(0) \gets $ Receive global model from the server
            \State $s \gets 1$
            \State $\mathcal{L}_\textit{history} \gets \emptyset$
            \State $\hat{s} \gets \infty,\; \hat{p} \gets 0$
            \While{true}
                \State $l_\textit{batch} \gets l\left(\theta(s-1);\,\mathcal{B}_s\right)$
                \State $\theta(s) \gets \theta(s-1) -\eta\nabla l_\textit{batch}$
                \State Add $(s, l_\textit{batch})$ to $\mathcal{L}_\textit{history}$
                \If{$s \ge \max(1,s_0-1)$ \textbf{and} $\hat{s}=\infty$}
                    \State $\hat{s}, \hat{p} \gets \Call{TALLOP}{\mathcal{L}_\textit{history}}$
                \EndIf
                \If{$s+1 \ge \hat{s}$}
                    \State Transmit $\mathcal{S}(\theta(\hat{s}) - \theta;\;\hat{p})$ to the server
                    \State \Return
                \EndIf
                \State $s \gets s + 1$
            \EndWhile
        \EndFunction
    \end{algorithmic}
\end{algorithm}

\section{Experiments and Discussion}

In this section, we examine the performance of TALLOP. We first compare its time-to-accuracy performance with three baseline methods and three ablations of TALLOP. We then analyze the computational complexity of TALLOP.

\begin{table}
    \centering
    \caption{Distribution of client computation speeds and communication bandwidths in the experiments. $m$ and $\sigma$ represent the mean and log-standard deviation of the log-normal distribution, respectively.} 
    \label{tab:client-config}
    \setlength{\tabcolsep}{6pt}
    \renewcommand{\arraystretch}{1.18}
    \begin{tabular}{lccccccccc}
        \toprule
        & \multicolumn{4}{c}{Computation [GFLOPS]} & \multicolumn{4}{c}{Communication [Mbps]} \\
        \cmidrule{2-5} \cmidrule{6-9} 
        & $m$ & $\sigma$ & min & max & $m$ & $\sigma$ & min & max \\
        \midrule
        Config. 1 & 20 & 0.5 & 5.48 & 61.64 & 80 & 1.0 & 3.83 & 491.25 \\
        Config. 2 & 200 & 0.5 & 54.8 & 616.4 & 80 & 1.0 & 3.83 & 491.25 \\
        \bottomrule
    \end{tabular}
\end{table}

\subsection{Time-to-Accuracy Evaluation}

First, we empirically compare TALLOP with existing baseline methods and its ablations.

We evaluate three dataset/model combinations: MobileNetV2~\cite{sandler_mobilenetv2_2018} on CIFAR-10~\cite{cifar10}, ViT~\cite{dosovitskiy_image_2020} on CIFAR-10, and ViT on CIFAR-100~\cite{cifar10}. For each dataset, the federation consists of 100 clients with non-IID data generated by a Dirichlet distribution ($\alpha=0.5$), and 20 clients are randomly selected in each communication round. 

We evaluate heterogeneous environments through simulations based on computation and communication performance models rather than physical experiments. Client computation and communication times are determined by their computational capability and communication bandwidth, respectively, which are independently sampled from log-normal distributions for each client. We consider two configurations: Configuration 1 has relatively low computational capability, while Configuration 2 has relatively high computational capability, making communication the dominant bottleneck. The corresponding distribution parameters are summarized in Table~\ref{tab:client-config}.

We compare TALLOP with three baseline methods: FFL~\cite{nori_fast_2021}, TimelyFL~\cite{zhang_timelyfl_2023}, and FedBalancer~\cite{shin_fedbalancer_2022}. We also consider three ablations to isolate the effects of the round deadline, adaptive local training, and communication compression:

\begin{enumerate}
\item[\textbf{(i)}] \textbf{Vanilla FL w/o deadline}:
All clients perform one local epoch in every round and transmit uncompressed model updates. The round duration is determined by the slowest participating client.

\item[\textbf{(ii)}] \textbf{Vanilla FL w/ deadline}:
All clients perform one local epoch and transmit uncompressed model updates. Updates arriving after the round deadline are discarded.

\item[\textbf{(iii)}] \textbf{Training-only optimization}:
Each client adapts its local training workload to maximize learning progress within the round deadline while always transmitting uncompressed model updates.

\end{enumerate}

TALLOP and the deadline-based ablations use per-round deadlines of 50 and 25 seconds under Configurations 1 and 2, respectively. The global wall-clock training limits for CIFAR-10 and CIFAR-100 are 20,000 and 40,000 seconds under Configuration 1, and 2,500 and 5,000 seconds under Configuration 2, respectively. Final accuracy is measured at the global training limit, whereas TTA is the first wall-clock time at which the evaluated global model reaches the specified target accuracy.

FFL, TimelyFL, and FedBalancer are evaluated only on MobileNetV2/CIFAR-10 under both system configurations, using the same data-heterogeneity and system settings as TALLOP. The three ablations are evaluated for all dataset/model combinations.

\begin{figure}[t]
    \centering
    \includegraphics[width=0.7\linewidth]{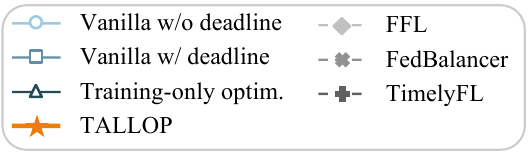}
    \begin{subfigure}{\linewidth}
        \centering
        \includegraphics[width=\linewidth]{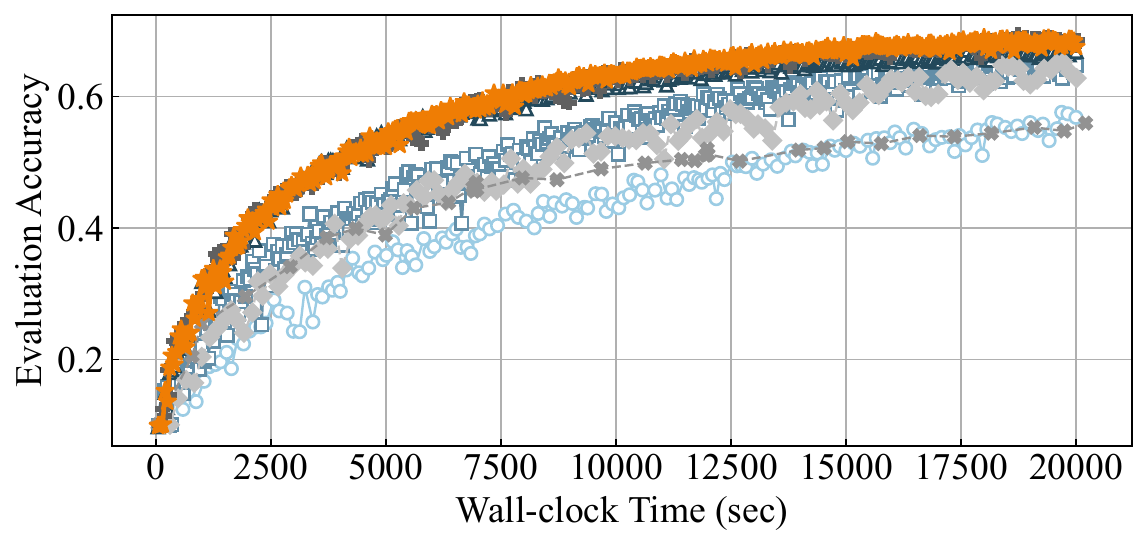}
        \caption{Configuration 1}
    \end{subfigure}
    \begin{subfigure}{\linewidth}
        \centering
        \includegraphics[width=\linewidth]{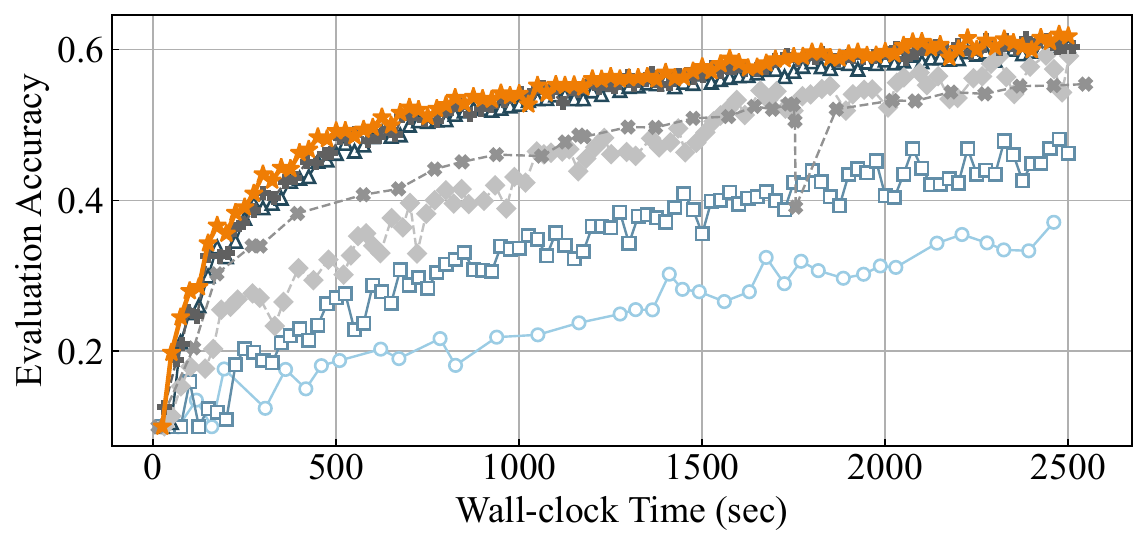}
        \caption{Configuration 2}
    \end{subfigure}
    \caption{Test accuracy as a function of wall-clock time for MobileNetV2 on CIFAR-10 under the two system configurations.}
    \label{fig:time-to-acc-result}
\end{figure}

\begin{table*}[t]
    \centering
    \caption{Final accuracy (FA, \%) and time-to-accuracy (TTA, seconds), rounded to the nearest second. ``--'' indicates that the target was not reached. FFL, TimelyFL, and FedBalancer are evaluated for MobileNetV2/CIFAR-10.}
    \label{tab:time_to_accuracy_results}
    \small
    \setlength{\tabcolsep}{2.35pt}
    \renewcommand{\arraystretch}{0.96}
    \begin{tabular}{ll@{\hspace{12pt}}r@{\hspace{6pt}}rrrr@{\hspace{16pt}}r@{\hspace{6pt}}rrrr}
        \toprule
        \multirow{2}{*}{\textbf{Dataset/model}} & \multirow{2}{*}{\textbf{Method}}
        & \multicolumn{5}{c}{\textbf{Config. 1 (slow comput.)}} & \multicolumn{5}{c}{\textbf{Config. 2 (fast comput.)}} \\
        \cmidrule(lr){3-7}\cmidrule(lr){8-12}
        & & \multicolumn{1}{c}{FA} & \multicolumn{4}{c}{TTA at targets} & \multicolumn{1}{c}{FA} & \multicolumn{4}{c}{TTA at targets} \\
        & & & 30\% & 40\% & 50\% & 60\% & & 30\% & 40\% & 50\% & 60\% \\
        \midrule
        \multirow{7}{*}{\shortstack[l]{CIFAR-10\\MobileNetV2}}
        & Vanilla w/o deadline
            & 56.83 & 3239 & 7140 & 12532 & --    
            & 37.10 & 1410 & --   & --   & -- \\
        & Vanilla w/ deadline
            & 64.75 & 1600 & 3350 & 6400  & 11900 
            & 46.25 & 675  & 1450 & --   & -- \\
        & Training-only optim. 
            & 66.85 & \textbf{1000} & 2000 & \textbf{3600} & 8150 
            & 60.53 & 175 & 350 & 725 & 2325 \\
        \addlinespace[3pt]
        & FFL 
            & 62.79 & 2214 & 3886 & 7718 & 14011 
            & 59.14 & 397  & 769  & 1531 & -- \\
        & TimelyFL
            & \textbf{68.55} & 1100 & 2010 & 3855 & 7680 
            & 60.34 & \textbf{149} & 301 & 679 & \textbf{2031} \\
        & FedBalancer 
            & 55.95 & 2921 & 5611 & 11426 & -- 
            & 55.42 & 174  & 574 & 1475 & -- \\
        \addlinespace[3pt]
        & TALLOP 
            & 67.55 & \textbf{1000} & \textbf{1950} & 4100 & \textbf{7350} 
            & \textbf{61.77} & 150 & \textbf{275} & \textbf{625} & 2050 \\
        \midrule
        \multirow{4}{*}{\shortstack[l]{CIFAR-10\\ViT}}
        & Vanilla w/o deadline 
            & 41.57 & 4423 & 17225 & -- & -- 
            & 23.51 & --   & --   & --   & -- \\
        & Vanilla w/ deadline 
            & 49.63 & 2550 & 8500  & 19350 & -- 
            & 31.37 & 1450 & --   & --   & -- \\
        & Training-only optim. 
            & \textbf{55.92} & \textbf{1000} & 3000 & 6150 & -- 
            & 51.09 & 250 & 625 & 1800 & -- \\
        \addlinespace[3pt]
        & TALLOP 
            & 55.58 & \textbf{1000} & \textbf{2950} & \textbf{5950} & -- 
            & \textbf{52.31} & \textbf{225} & \textbf{525} & \textbf{1400} & -- \\
        \bottomrule
    \end{tabular}

    \vspace{6pt}
    \begin{tabular}{ll@{\hspace{12pt}}r@{\hspace{6pt}}rrr@{\hspace{16pt}}r@{\hspace{6pt}}rrr}
        \toprule
        \multirow{2}{*}{\textbf{Dataset/model}} & \multirow{2}{*}{\textbf{Method}}
        & \multicolumn{4}{c}{\textbf{Config. 1 (slow comput.)}} & \multicolumn{4}{c}{\textbf{Config. 2 (fast comput.)}} \\
        \cmidrule(lr){3-6}\cmidrule(lr){7-10}
        & & \multicolumn{1}{c}{FA} & \multicolumn{3}{c}{TTA at targets} & \multicolumn{1}{c}{FA} & \multicolumn{3}{c}{TTA at targets} \\
        & & & 10\% & 20\% & 30\% & & 10\% & 20\% & 30\% \\
        \midrule
        \multirow{4}{*}{\shortstack[l]{CIFAR-100\\ViT}}
        & Vanilla w/o deadline 
            & 24.96 & 8864 & 26232 & -- & 8.08 & -- & -- & -- \\
        & Vanilla w/ deadline 
            & \textbf{32.59} & 3850 & 11350 & 26550 & 18.08 & 1850 & -- & -- \\
        & Training-only optim.
            & 28.92 & 1200 & 3100 & 14300 & 27.45 & 275 & 800 & -- \\
        \addlinespace[3pt]
        & TALLOP 
            & 29.38 & \textbf{1100} & \textbf{2950} & \textbf{13300} & \textbf{28.26} & \textbf{250} & \textbf{675} & -- \\
        \bottomrule
    \end{tabular}
\end{table*}

Figure~\ref{fig:time-to-acc-result} shows the wall-clock accuracy trajectories for MobileNetV2 on CIFAR-10, while Table~\ref{tab:time_to_accuracy_results} summarizes the final accuracy (FA) and time-to-accuracy (TTA) across all evaluated dataset/model combinations. Under Configuration~2, where computation is relatively fast, TALLOP achieves the highest FA across all evaluated model-dataset combinations. It also achieves the shortest TTA at most accuracy targets, except for the two targets where TimelyFL outperforms it. In contrast, under Configuration~1, where computation is slower, TALLOP's advantage in FA is less pronounced, with other baselines or ablations achieving higher FA in several settings. Nevertheless, TALLOP maintains the shortest or tied-shortest TTA at most targets. These results indicate that TALLOP generally reduces the wall-clock time required to reach a given accuracy, although its benefit in FA depends on the system configuration.

The ablation results further highlight the benefits of deadline-based adaptive FL. Compared with vanilla FL, introducing a deadline generally improves both FA and TTA, reducing TTA by half or more at several targets. Adapting the training workload further improves TTA in most settings and FA in many settings. TALLOP, which additionally adapts communication compression, achieves shorter TTA than training-only optimization at most targets. However, the improvements are relatively small in several settings, indicating that the additional benefit of joint computation and communication optimization is not always substantial, particularly when computation is the dominant bottleneck.

These experiments rely on computation and communication performance models rather than execution on a real distributed system. Consequently, they do not capture deployment effects beyond the models, such as runtime variability, communication contention, or protocol overhead. Evaluating TALLOP using measurements from real deployments or empirical system traces remains important future work. Moreover, the experiments were conducted with a single random seed, without error bars; additional runs are needed to assess the statistical robustness of the observed improvements.

\subsection{Computational Overhead Analysis of TALLOP}

Next, we analyze the additional computational overhead introduced by TALLOP based on Algorithm~\ref{alg:proposal} in terms of the number of operations. 

In TALLOP, the time allocation factor $\rho \in [0,1]$ is treated as the optimization variable, and $\Call{PredictCompressedLoss}{\rho}$ is evaluated during ternary search. Let $\varepsilon$ denote the termination threshold for the search interval. Then, the number of evaluations of $\Call{PredictCompressedLoss}{\rho}$ is $O(\log(1/\varepsilon))$.

Next, we estimate the computational complexity of a single call to $\Call{PredictCompressedLoss}{\rho}$. The dominant operations are (i) generating the sparsified model parameters $\theta(s_0, p(\rho))$ and (ii) computing the $\ell_2$ distance to the original parameters $\theta$. Let $P = |\theta|$ denote the total number of model parameters. Operation (i) requires selecting or sorting parameters according to their magnitudes, resulting in a computational complexity of $O(P \log P)$. Operation (ii) requires scanning all model parameters to compute the Euclidean distance, resulting in a computational complexity of $O(P)$.

Therefore, the overall additional computational complexity of TALLOP is estimated as $O(P \log P \, \log (1/\varepsilon))$ per communication round. As a concrete example, consider MobileNetV2, which has approximately $P \approx 2 \times 10^6$ parameters. Assuming $\log (1/\varepsilon) \approx 10$, the additional computation introduced by TALLOP is on the order of approximately $10^8$ to $10^9$ operations per communication round. In contrast, training MobileNetV2 requires approximately $9 \times 10^8$ floating-point multiply-add operations per training sample. Therefore, the additional computation introduced by TALLOP corresponds to, at most, the computational cost of training on only a few samples. Since federated learning typically performs local training on many samples in each communication round, this additional overhead has only a negligible impact on the overall training time.

\section{Conclusion}

In this paper, we addressed time-to-accuracy optimization in FL under heterogeneous environments where communication time is non-negligible. We proposed TALLOP, which jointly optimizes the local training workload and communication compression ratio under a fixed per-round time budget by predicting the post-compression model loss for different training-communication time allocations. In the evaluated settings, TALLOP generally matched or improved TTA relative to training-only optimization and was competitive with the three baseline methods on MobileNetV2/CIFAR-10. However, the improvement was modest or absent in some settings, and the baseline comparison was limited to a single model and random seed.

This work assumes that the round time budget is predefined. An important direction for future work is to jointly optimize the round time budget and the training-communication time allocation. Another direction is to evaluate TALLOP under more realistic deployment scenarios with dynamically changing computation capabilities, fluctuating communication bandwidths, and network congestion, as well as to extend TALLOP to asynchronous federated learning.

\bibliography{main}
\bibliographystyle{ieeetr}

\end{document}